# Planning Graph as a (Dynamic) CSP:
# Exploiting EBL, DDB and other CSP Search Techniques in Graphplan

**Subbarao Kambhampati**                                                    RAO@ASU.EDU
*Department of Computer Science and Engineering*
*Arizona State University, Tempe AZ 85287-5406*

## Abstract

This paper reviews the connections between Graphplan's planning-graph and the dynamic constraint satisfaction problem and motivates the need for adapting CSP search techniques to the Graphplan algorithm. It then describes how explanation based learning, dependency directed backtracking, dynamic variable ordering, forward checking, sticky values and random-restart search strategies can be adapted to Graphplan. Empirical results are provided to demonstrate that these augmentations improve Graphplan's performance significantly (up to 1000x speedups)on several benchmark problems. Special attention is paid to the explanation-based learning and dependency directed backtracking techniques as they are empirically found to be most useful in improving the performance of Graphplan.

## 1. Introduction

Graphplan (Blum & Furst, 1997) is currently one of the more efficient algorithms for solving classical planning problems. Four of the five competing systems in the recent AIPS-98 planning competition were based on the Graphplan algorithm (McDermott, 1998). Extending the efficiency of the Graphplan algorithm thus seems to be a worth-while activity. In (Kambhampati, Parker, & Lambrecht, 1997), we provided a reconstruction of Graphplan algorithm to explicate its links to previous work in classical planning and constraint satisfaction. One specific link that was discussed is the connection between the process of searching Graphplan's planning graph, and solving a "dynamic constraint satisfaction problem" (DCSP) (Mittal & Falkenhainer, 1990). Seen from the DCSP perspective, the standard backward search proposed by Blum and Furst (1997) lacks a variety of ingredients that are thought to make up efficient CSP search mechanisms (Frost & Dechter, 1994; Bayardo & Schrag, 1997). These include forward checking, dynamic variable ordering, dependency directed backtracking and explanation-based learning (Tsang, 1993; Kambhampati, 1998). In (Kambhampati et al., 1997), I have suggested that it would be beneficial to study the impact of these extensions on the effectiveness of Graphplan's backward search.

In this paper, I describe my experiences with adding a variety of CSP search techniques to improve Graphplan backward search–including explanation-based learning (EBL) and dependency-directed backtracking capabilities (DDB), Dynamic variable ordering, Forward checking, sticky values, and random-restart search strategies. Of these, the addition of EBL and DDB capabilities turned out to be empirically the most useful. Both EBL and DDB are based on explaining failures at the leaf-nodes of a search tree, and propagating those explanations upwards through the search tree (Kambhampati, 1998). DDB involves using the propagation of failure explanations to support intelligent backtracking, while EBL involves storing interior-node failure explanations, for pruning future search nodes. Graphplan does use a weak form of failure-driven learning that it calls "mem-oization." As we shall see in this paper, Graphplan's brand of learning is quite limited as there is no explicit analysis of the reasons for failure. Instead the explanation of failure of a search node is taken to be *all* the constraints in that search node. As explained in (Kambhampati, 1998), this not only eliminates the opportunities for dependency directed backtracking, it also adversely effects the utility of the stored memos.

Adding full-fledged EBL and DDB capabilities in effect gives Graphplan both the ability to do intelligent backtracking, and the ability to learn generalized memos that are more likely to be applicable in other situations. Technically, this involves generalizing conflict-directed backjumping (Prosser, 1993), a specialized version of EBL/DDB strategy applicable for binary CSP problems[1] to work in the context of dynamic constraint satisfaction problems (as the usual bane of no-good learning strategies–under control, it also loses some learning opportunities. I then present the use Empirically, the EBL/DDB capabilities improve Graphplan's search efficiency quite dramatically–giving rise to up to 1000x speedups, and allowing Graphplan to easily solve several problems that have hither-to been hard or unsolvable. In particular, I will report on my experiments with the bench-mark problems described by Kautz and Selman (1996), as well as 4 other domains, some of which were used in the recent AIPS planning competition (McDermott, 1998).

I discuss the utility issues involved in storing and using memos, and point out that the Graphplan memoization strategy can be seen as a *very* conservative form of CSP no-good learning. While this conservative strategy keeps the storage and retrieval costs of no-goods –the usual bane of no-good learning strategies–under control, it also loses some learning opportunities. I then present the use of "sticky values" as a way of recouping some of these losses. Empirical studies show that sticky values lead to a further 2-4x improvement over EBL.

In addition to EBL and DDB, I also investigated the utility of forward checking and dynamic variable ordering, both in isolation and in concert with EBL and DDB. My empirical studies show that these capabilities typically lead to an additional 2-4x speedup over EBL/DDB, but are not by themselves competitive with EBL/DDB.

Finally, I consider the utility of the EBL/DDB strategies in the context of random-restart search strategies (Gomes, Selman, & Kautz, 1998) that have recently been shown to be good at solving hard combinatorial problems, including planning problems. My results show that EBL/DDB strategies retain their advantages even in the context of such random-restart strategies. Specifically, EBL/DDB strategies enable Graphplan to use the backtrack limits more effectively–allowing it to achieve higher solvability rates, and more optimal plans with significantly smaller backtrack and restart limits.

This paper is organized as follows. In the next section, I provide some background on viewing Graphplan's backward search as a (dynamic) constraint satisfaction problem, and review some of the opportunities this view presents. In Section 3, I discuss some inefficiencies of the backtracking and learning methods used in normal Graphplan that motivate the need for EBL/DDB capabilities. Section 4 describes how EBL and DDB are added to Graphplan. Section 5 presents empirical studies demonstrating the usefulness of these augmentations. Section 7 investigates the utility of forward checking and dynamic variable ordering strategies for Graphplan. Section 8 investigates the utility of EBL/DDB strategies in the context of random-restart search. Section 9 discusses related work and Section 10 presents conclusions and some directions for further work.

---

1. Binary CSP problems are those problems where all initial constraints are between pairs of variables.

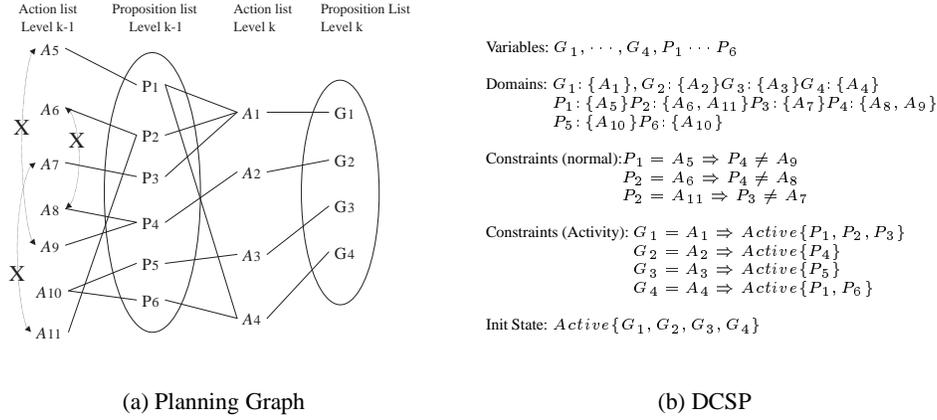

(a) Planning Graph                    (b) DCSP

Figure 1: A planning graph and the DCSP corresponding to it

## 2. Review of Graphplan Algorithm and its Connections to DCSP

### 2.1 Review of Graphplan Algorithm

Graphplan algorithm (Blum & Furst, 1997) can be seen as a "disjunctive" version of the forward state space planners (Kambhampati et al., 1997; Kambhampati, 1997). It consists of two interleaved phases – a forward phase, where a data structure called "planning-graph" is incrementally extended, and a backward phase where the planning-graph is searched to extract a valid plan. The planning-graph consists of two alternating structures, called proposition lists and action lists. Figure 1 shows a partial planning-graph structure. We start with the initial state as the zeroth level proposition list. Given a $k$ level planning graph, the extension of structure to level $k + 1$ involves introducing all actions whose preconditions are present in the $k^{th}$ level proposition list. In addition to the actions given in the domain model, we consider a set of dummy "persist" actions, one for each condition in the $k^{th}$ level proposition list. A "persist-C" action has $C$ as its precondition and $C$ as its effect. Once the actions are introduced, the proposition list at level $k + 1$ is constructed as just the union of the effects of all the introduced actions. Planning-graph maintains the dependency links between the actions at level $k+1$ and their preconditions in level $k$ proposition list and their effects in level $k + 1$ proposition list. The planning-graph construction also involves computation and propagation of "mutex" constraints. The propagation starts at level 1, with the actions that are statically interfering with each other (i.e., their preconditions and effects are inconsistent) labeled mutex. Mutexes are then propagated from this level forward by using a two simple rules: two propositions at level $k$ are marked mutex if all actions at level $k$ that support one proposition are mutex with all actions that support the second proposition. Two actions at level $k + 1$ are mutex if they are statically interfering or if one of the propositions (preconditions) supporting the first action is mutually exclusive with one of the propositions supporting the second action.

The search phase on a $k$ level planning-graph involves checking to see if there is a sub-graph of the planning-graph that corresponds to a valid solution to the problem. This involves starting with the propositions corresponding to goals at level $k$ (if all the goals are not present, or if they are present but a pair of them are marked mutually exclusive, the search is abandoned right away, and planning-grap is grown another level). For each of the goal propositions, we then select an action

Variables: $G_1, \cdots, G_4, P_1 \cdots P_6$

Domains: $G_1$: $\{A_1\}, G_2$: $\{A_2\} G_3$: $\{A_3\} G_4$: $\{A_4\}$
$\quad\quad P_1$: $\{A_5\} P_2$: $\{A_6, A_{11}\} P_3$: $\{A_7\} P_4$: $\{A_8, A_9\}$
$\quad\quad P_5$: $\{A_{10}\} P_6$: $\{A_{10}\}$

Constraints (normal): $P_1 = A_5 \Rightarrow P_4 \neq A_9$
$\quad\quad\quad\quad\quad P_2 = A_6 \Rightarrow P_4 \neq A_8$
$\quad\quad\quad\quad\quad P_2 = A_{11} \Rightarrow P_3 \neq A_7$

Constraints (Activity): $G_1 = A_1 \Rightarrow Active\{P_1, P_2, P_3\}$
$\quad\quad\quad\quad\quad\quad G_2 = A_2 \Rightarrow Active\{P_4\}$
$\quad\quad\quad\quad\quad\quad G_3 = A_3 \Rightarrow Active\{P_5\}$
$\quad\quad\quad\quad\quad\quad G_4 = A_4 \Rightarrow Active\{P_1, P_6\}$

Init State: $Active\{G_1, G_2, G_3, G_4\}$

**(a) DCSP**

Variables: $G_1, \cdots, G_4, P_1 \cdots P_6$

Domains: $G_1$: $\{A_1, \perp\}, G_2$: $\{A_2, \perp\} G_3$: $\{A_3, \perp\} G_4$: $\{A_4, \perp\}$
$\quad\quad P_1$: $\{A_5, \perp\} P_2$: $\{A_6, A_{11}, \perp\} P_3$: $\{A_7, \perp\} P_4$: $\{A_8, A_9, \perp\}$
$\quad\quad P_5$: $\{A_{10}, \perp\} P_6$: $\{A_{10}, \perp\}$

Constraints (normal): $P_1 = A_5 \Rightarrow P_4 \neq A_9$
$\quad\quad\quad\quad\quad P_2 = A_6 \Rightarrow P_4 \neq A_8$
$\quad\quad\quad\quad\quad P_2 = A_{11} \Rightarrow P_3 \neq A_7$

Constraints (Activity): $G_1 = A_1 \Rightarrow P_1 \neq \perp \wedge P_2 \neq \perp \wedge P_3 \neq \perp$
$\quad\quad\quad\quad\quad\quad G_2 = A_2 \Rightarrow P_4 \neq \perp$
$\quad\quad\quad\quad\quad\quad G_3 = A_3 \Rightarrow P_5 \neq \perp$
$\quad\quad\quad\quad\quad\quad G_4 = A_4 \Rightarrow P_1 \neq \perp \wedge P_6 \neq \perp$

Init State: $G_1 \neq \perp \wedge G_2 \neq \perp \wedge G_3 \neq \perp \wedge G_4 \neq \perp$

**(b) CSP**

Figure 2: Compiling a DCSP to a standard CSP

from the level $k$ action list that supports it, such that no two actions selected for supporting two different goals are mutually exclusive (if they are, we backtrack and try to change the selection of actions). At this point, we recursively call the same search process on the $k-1$ level planning-graph, with the preconditions of the actions selected at level $k$ as the goals for the $k-1$ level search. The search succeeds when we reach level 0 (corresponding to the initial state).

Consider the (partial) planning graph shown in Figure 3 that Graphplan may have generated and is about to search for a solution. $G_1 \cdots G_4$ are the top level goals that we want to satisfy, and $A_1 \cdots A_4$ are the actions that support these goals in the planning graph. The specific action-precondition dependencies are shown by the straight line connections. The actions $A_5 \cdots A_{11}$ at the left-most level support the conditions $P_1 \cdots P_6$ in the planning-graph. Notice that the conditions $P_2$ and $P_4$ at level $k-1$ are supported by two actions each. The x-marked connections between the actions $A_5, A_9$, $A_6, A_8$ and $A_7, A_{11}$ denote that those action pairs are mutually exclusive. (Notice that given these mutually exclusive relations alone, Graphplan cannot derive any mutual exclusion relations at the proposition level $P_1 \cdots P_6$.)

## 2.2 Connections Between Graphplan and CSP

The Graphplan algorithm as described above bears little resemblance to previous classical planning algorithms. In (Kambhampati et al., 1997), we explicate a number of important links between the Graphplan algorithm and previous work in planning and constraint satisfaction communities. Specifically, I show that a planning-graph of length $k$ can be thought of (to a first approximation) as a disjunctive (unioned) version of a $k$-level search tree generated by a forward state-space refinement, with the action lists corresponding to the union of all actions appearing at $k^{th}$ level, and proposition lists corresponding to the union of all states appearing at the $k^{th}$ level. The mutex constraints can be seen as providing (partial) information about which subsets of a proposition list actually correspond to legal states in the corresponding forward state-space search. The process of searching the planning graph to extract a valid plan from it can be seen as a dynamic constraint satisfaction problem. Since this last link is most relevant to the work described in this paper, I will review it further below.

The dynamic constraint satisfaction problem (DCSP) (Mittal & Falkenhainer, 1990) is a generalization of the constraint satisfaction problem (Tsang, 1993), that is specified by a set of variables, activity flags for the variables, the domains of the variables, and the constraints on the legal variable-value combinations. In a DCSP, initially only a subset of the variables is active, and the objective is to find assignments for all active variables that is consistent with the constraints among those variables. In addition, the DCSP specification also contains a set of "activity constraints." An activity constraint is of the form: "if variable $x$ takes on the value $v_x$, then the variables $y, z, w...$ become active."

The correspondence between the planning-graph and the DCSP should now be clear. Specifically, the propositions at various levels correspond to the DCSP variables[2], and the actions supporting them correspond to the variable domains. There are three types of constraints: *action mutex constraints, fact (proposition) mutex constraints* and *subgoal activation constraints*.

Since actions are modeled as values rather than variables, action mutex constraints have to be modeled indirectly as constraints between propositions. If two actions $a_1$ and $a_2$ are marked mutex with each other in the planning graph, then for *every pair* of propositions $p_{11}$ and $p_{12}$ where $a_1$ is one of the possible supporting actions for $p_{11}$ and $a_2$ is one of the possible supporting actions for $p_{12}$, we have the constraint:

$$\neg (p_{11} = a_1 \wedge p_{12} = a_2)$$

Fact mutex constraints are modeled as constraints that prohibit the simultaneous activation of the two facts. Specifically, if two propositions $p_{11}$ and $p_{12}$ are marked mutex in the planning graph, we have the constraint:

$$\neg (Active(p_{11}) \wedge Active(p_{12}))$$

Subgoal activation constraints are implicitly specified by action preconditions: supporting an active proposition $p$ with an action $a$ makes all the propositions in the previous level corresponding to the preconditions of $a$ active.

Finally, only the propositions corresponding to the goals of the problem are "active" in the beginning. Figure 1 shows the dynamic constraint satisfaction problem corresponding to the example planning-graph that we discussed.

### 2.2.1 SOLVING A DCSP

There are two ways of solving a DCSP problem. The first, direct, approach (Mittal & Falkenhainer, 1990) involves starting with the initially active variables, and finding a satisfying assignment for them. This assignment may activate some new variables, and these newly activated variables are assigned in the second epoch. This process continues until we reach an epoch where no more new variables are activated (which implies success), or we are unable to give a satisfying assignment to the activated variables at a given epoch. In this latter case, we backtrack to the previous epoch and try to find an alternative satisfying assignment to those variables (backtracking further, if no other assignment is possible). The backward search process used by the Graphplan algorithm (Blum & Furst, 1997) can be seen as solving the DCSP corresponding to the planning graph in this direct fashion.

The second approach for solving a DCSP is to first compile it into a standard CSP, and use the standard CSP algorithms. This compilation process is quite straightforward and is illustrated in

---

2. Note that the same literal appearing in different levels corresponds to different DCSP variables. Thus, strictly speaking, a literal $p$ in the proposition list at level $i$ is converted into a DCSP variable $p_i$. To keep matters simple, the example in Figure 1 contains syntactically different literals in different levels of the graph.

Figure 2. The main idea is to introduce a new "null" value (denoted by "$\perp$") into the domains of each of the DCSP variables. We then model an inactive DCSP variable as a CSP variable which takes the value $\perp$. The constraint that a particular variable $P$ be active is modeled as $P \neq \perp$. Thus, activity constraint of the form

$$G_1 = A_1 \Rightarrow Active\{P_1, P_2, P_3\}$$

is compiled to the standard CSP constraint

$$G_1 = A_1 \Rightarrow P_1 \neq \perp \wedge P_2 \neq \perp \wedge P_3 \neq \perp$$

It is worth noting here that the activation constraints above are only concerned about ensuring that propositions that are preconditions of a selected action do take non-$\perp$ values. They thus allow for the possibility that propositions can become active (take non-$\perp$ values) even though they are strictly not supporting preconditions of any selected action. Although this can lead to inoptimal plans, the mutex constraints ensure that no unsound plans will be produced (Kautz & Selman, 1999). To avoid unnecessary activation of variables, we need to add constraints to the effect that unless one of the actions needing that variable as a precondition has been selected as the value for some variable in the earlier (higher) level, the variable must have $\perp$ value. Such constraints are typically going to have very high arity (as they wind up mentioning a large number of variables in the previous level), and may thus be harder to handle during search.

Finally, a mutex constraint between two propositions

$$\neg\left(Active(p_{11}) \wedge Active(p_{12})\right)$$

is compiled into

$$\neg\left(p_{11} \neq \perp \wedge p_{12} \neq \perp\right).$$

Since action mutex constraints are already in the standard CSP form, with this compilation, all the activity constraints are converted into standard constraints and thus the entire CSP is now a standard CSP. It can now be solved by any of the standard CSP search techniques (Tsang, 1993).[3]

The direct method has the advantage that it closely mirrors the Graphplan's planning graph structure and its backward search. Because of this, it is possible to implement the approach on the plan graph structure without explicitly representing all the constraints. Furthermore, as I will discuss in Section 6, there are some distinct advantages for adopting the DCSP view in implementing EBL/DDB on Graphplan. The compilation to CSP requires that plan graph be first converted into an extensional CSP. It does however allow the use of standard algorithms, as well as supports non-directional search (in that one does not have to follow the epoch-by-epoch approach in assigning variables).[4] Since my main aim is to illustrate the utility of CSP search techniques in the context of the Graphplan algorithm, I will adopt the direct solution method for the DCSP. For a study of the tradeoffs offered by the technique of compiling the planning graph into a CSP, the reader is referred to (Do & Kambhampati, 2000).

---

3. It is also possible to compile any CSP problem to a propositional satisfiability problem (i.e., a CSP problem with boolean variables). This is accomplished by compiling every CSP variable $P$ that has the domain $\{v_1, v_2, \cdots, v_n\}$ into $n$ boolean variables of the form *P-is-$v_1$* $\cdots$ *P-is-$v_n$*. Every constraint of the form $P = v_j \wedge \cdots \Rightarrow \cdots$ is compiled to *P-is-$v_j$* $\wedge \cdots \Rightarrow \cdots$. This is essentially what is done by the BLACKBOX system (Kautz & Selman, 1999).

4. Compilation to CSP is not a strict requirement for doing non-directional search. In (Zimmerman & Kambhampati, 1999), we describe a technique that allows the backward search of Graphplan to be non-directional, see the discussion in Section 10.

## 2.3 Interpreting Mutex Propagation from the CSP View

Viewing the planning graph as a constraint satisfaction problem helps put the mutex propagation in a clearer perspective (see (Kambhampati et al., 1997)). Specifically, the way Graphplan constructs its planning graph, it winds up enforcing partial directed 1-consistency and partial directed 2-consistency (Tsang, 1993). The partial 1-consistency is ensured by the graph building procedure which introduces an action at level $l$ only if the actions preconditions are present in the proposition list at level $l - 1$ and are not mutually exclusive. Partial 2-consistency is ensured by the mutual exclusion propagation procedure.

In particular, the Graphplan planning graph construction implicitly derives "no-good"[5] constraints of the form:

$$\neg Active(P_m^i) \quad (or \ \ P_m^i \neq \bot)$$

In which case $P_m^i$ will be simply removed from (or will not be put into) the level $i$, and the mutex constraints of the form:

$$\neg \left( Active(P_m^i) \wedge Active(P_n^i) \right) \quad (or \ \ P_m^i \neq \bot \wedge P_n^i \neq \bot)$$

in which case $P_m^i$ and $P_n^i$ are marked mutually exclusive.

Both procedures are "directed" in that they only use "reachability" analysis in enforcing the consistency, and are "partial" in that they do not enforce either full 1-consistency or full 2-consistency. Lack of full 1-consistency is verified by the fact that the appearance of a goal at level $k$ does not necessarily mean that the goal is actually achievable by level $k$ (i.e., there is a solution for the CSP that assigns a non-$\bot$ value to that goal at that level). Similarly, lack of full 2-consistency is verified by the fact that appearance of a pair of goals at level $k$ does not imply that there is a plan for achieving both goals by that level.

There is another, somewhat less obvious, way in which the consistency enforcement used in Graphplan is partial (and very conservative)–it concentrates only on whether a single goal variable or a pair of goal variables can simultaneously have non-$\bot$ values (be active) in a solution. It may be that a goal can have a non-$\bot$ value, but not all non-$\bot$ values are feasible. Similarly, it may be that a pair of goals are achievable, but not necessarily achievable with every possible pair of actions in their respective domains.

This interpretation of mutex propagation procedure in Graphplan brings to fore several possible extensions worth considering for Graphplan:

1. Explore the utility of directional consistency enforcement procedures that are not based solely on reachability analysis. Kambhampati *et. al.* (1997) argue for extending this analysis using relevance information, and Do *et. al.* (2000) provide an empirical analysis of the effectiveness of consistency enforcement through relevance information.

2. Explore the utility of enforcing higher level consistency. As pointed out in (Kambhampati et al., 1997; Kambhampati, 1998), the memoization strategies can be seen as failure-driven procedures that incrementally enforce partial higher level consistency.

---

5. Normally, in the CSP literature, a no-good is seen as a compound assignment that can not be part of any feasible solution. With this view, mutex constraints of the form $P_m^i \neq \bot \wedge P_n^i \neq \bot$ correspond to a conjunction of nogoods of the the form $P_m^i = a_u \wedge P_n^i = a_v$ where $a_u$ and $a_v$ are values in the domains of $P_m^i$ and $P_n^i$.

3. Consider relaxing the focus on non- $\perp$ values alone, and allow derivation of no-goods of the form

$$P_m^i = a_u \wedge P_n^i = a_v$$

This is not guaranteed to be a winning idea as the number of derived no-goods can increase quite dramatically. In particular, assuming that there are $l$ levels in the planning graph, and an average of $m$ goals per level, and an average of $d$ actions supporting each goal, the maximum number of Graphplan style pair-wise mutexes will be $O(l * m^2)$ while the 2-size no-goods of type discussed here will be $O(l * (m * (d + 1))^2)$. We consider a similar issue in the context of Graphplan memoization strategy in Section 6.

## 3. Some Inefficiencies of Graphplan's Backward Search

To motivate the need for EBL and DDB, we shall first review the details of Graphplan's backward search, and pinpoint some of its inefficiencies. We shall base our discussion on the example planning graph from Figure 3 (which is reproduced for convenience from Figure 1). Assuming that $G_1 \cdots G_4$ are the top level goals of the problem we are interested in solving, we start at level $k$, and select actions to support the goals $G_1 \cdots G_4$. To keep matters simple, we shall assume that the search assigns the conditions (variables) at each level from top to bottom (i.e., $G_1$ first, then $G_2$ and so on). Further, when there is a choice in the actions (values) that can support a condition, we will consider the top actions first. Since there is only one choice for each of the conditions at this level, and none of the actions are mutually exclusive with each other, we select the actions $A_1, A_2, A_3$ and $A_4$ for supporting the conditions at level $k$. We now have to make sure that the preconditions of $A_1, A_2, A_3, A_4$ are satisfied at level $k - 1$. We thus subgoal on the conditions $P_1 \cdots P_6$ at level $k - 1$, and recursively start the action selection for them. We select the action $A_5$ for $P_1$. For $P_2$, we have two supporting actions, and using our convention, we select $A_6$ first. For $P_3$, $A_7$ is the only choice. When we get down to selecting a support for $P_4$, we again have a choice. Suppose we select $A_8$ first. We find that this choice is infeasible as $A_8$ is mutually exclusive with $A_6$ that is already chosen. So, we backtrack and choose $A_9$, and find that it too is mutually exclusive with a previously selected action, $A_5$. We now are stymied as there are no other choices for $P_4$. So, we have to backtrack and undo choices for the previous conditions. Graphplan uses a chronological backtracking approach, whereby, it first tries to see if $P_3$ can be re-assigned, and then $P_2$ and so on. Notice the **first** indication of inefficiency here – the failure to assign $P_4$ had nothing to do with the assignment for $P_3$, and yet, chronological backtracking will try to re-assign $P_3$ in the vain hope of averting the failure. This can lead to a large amount of wasted effort had it been the case that $P_3$ did indeed have other choices.

As it turns out, we find that $P_3$ has no other choices and backtrack over it. $P_2$ does have another choice – $A_{11}$. We try to continue the search forward with this value for $P_2$, but hit an impasse at $P_3$– since the only value of $P_3$, $A_7$ is mutex with $A_{11}$. At this point, we backtrack over $P_3$, and continue backtracking over $P_2$ and $P_1$, as they too have no other remaining choices. When we backtrack over $P_1$, we need to go back to level $k$ and try to re-assign the goals at that level. Before this is done, the Graphplan search algorithm makes a "memo" signifying the fact that it failed to satisfy the goals $P_1 \cdots P_6$ at this level, with the hope that if the search ever subgoals on these same set of goals in future, we can scuttle it right away with the help of the remembered memo. Here is the **second** indication of inefficiency – we are remembering all the subgoals $P_1 \cdots P_6$ even though we can see that the problem lies in trying to assign $P_1, P_2, P_3$ and $P_4$ simultaneously, and has nothing to do with the other subgoals. If we remember $\{P_1, P_2, P_3, P_4\}$ as the memo as against $\{P_1 \cdots P_6\}$, the remembered memo would be more general, and would have a much better chance of being useful in the future.

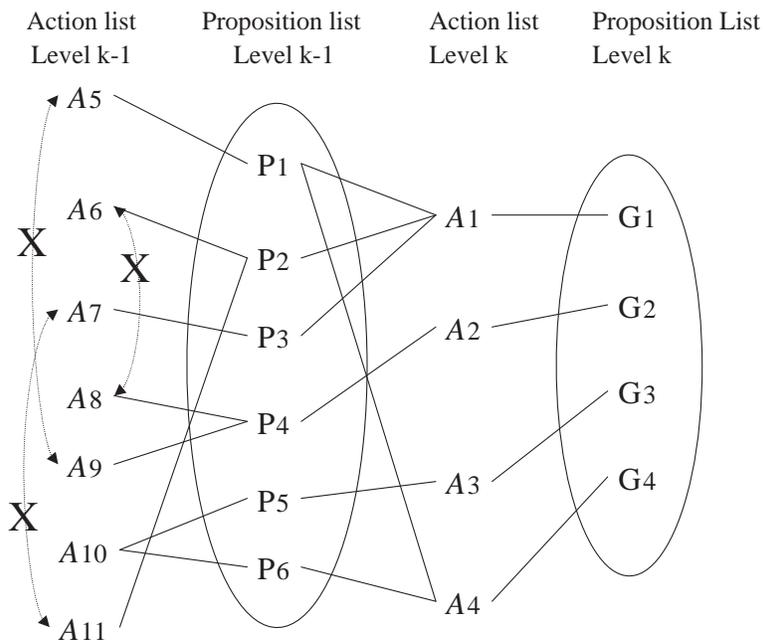

Figure 3: The running example used to illustrate EBL/DDB in Graphplan

After the memo is stored, the backtracking continues into level $k$ – once again in a chronological fashion, trying to reassign $G_4, G_3, G_2$ and $G_1$ in that order. Here we see the **third** indication of inefficiency caused by chronological backtracking – $G_3$ really has no role in the failure we encountered in assigning $P_3$ and $P_4$ – since it only spawns the condition $P_5$ at level $k - 1$. Yet, the backtracking scheme of Graphplan considers reassigning $G_3$. A somewhat more subtle point is that reassigning $G_4$ is not going to avert the failure either. Although $G_4$ requires $P_1$ one of the conditions taking part in the failure, $P_1$ is also required by $G_1$ and unless $G_1$ gets reassigned, considering further assignments to $G_4$ is not going to avert the failure.

For this example, we continue backtracking over $G_2$ and $G_1$ too, since they too have no alternative supports, and finally memoize $\{G_1, G_2, G_3, G_4\}$ at this level. At this point the backward search fails, and Graphplan extends the planning graph by another level before re-initiating the backward search on the extended graph.

## 4. Improving Backward Search with EBL and DDB

I will now describe how Graphplan's backward search can be augmented with full fledged EBL and DDB capabilities to eliminate the inefficiencies pointed out in the previous section. Informally, EBL/DDB strategies involve explanation of failures at leaf nodes, and regression and propagation of leaf node failure explanations to compute interior node failure explanations, along the lines described in (Kambhampati, 1998). The specific extensions I propose to the backward search can essentially be seen as adapting conflict-directed backjumping strategy (Prosser, 1993), and generalizing it to work with dynamic constraint satisfaction problems.

The algorithm is shown in pseudo-code form in Figure 4. It contains two mutually recursive procedures `find-plan` and `assign-goals`. The former is called once for each level of the planning-graph. It then calls `assign-goals` to assign values to all the required conditions at that level. `assign-goals` picks a condition, selects a value for it, and recursively calls itself with the remaining conditions. When it is invoked with empty set of conditions to be assigned, it calls find-plan to initiate the search at the next (previous) level.

In order to illustrate how EBL/DDB capabilities are added, let's retrace the previous example, and pick up at the point where we are about to assign $P_4$ at level $k - 1$, having assigned $P_1, P_2$ and $P_3$. When we try to assign the value $A_8$ to $P_4$, we violate the mutex constraint between $A_6$ and $A_8$. An explanation of failure for a search node is a set of constraints from which $False$ can be derived. The complete explanation for this failure can thus be stated as:

$$P_2 = A_6 \land P_4 = A_8 \land (P_2 = A_6 \Rightarrow P_4 \neq A_8)$$

Of this, the part $P_2 = A_6 \Rightarrow P_4 \neq A_8$ can be stripped from the explanation since the mutual exclusion relation will hold as long as we are solving this particular problem with these particular actions. Further, we can take a cue from the conflict directed backjumping algorithm (Prosser, 1993), and represent the remaining explanation compactly in terms of "conflict sets." Specifically, whenever the search reaches a condition $c$ (and is about to find an assignment for it), its conflict set is initialized as $\{c\}$. Whenever one of the possible assignments to $c$ is inconsistent (mutually exclusive) with the current assignment of a previous variable $c'$, we add $c'$ to the conflict set of $c$. In the current example, we start with $\{P_4\}$ as the conflict set of $P_4$, and expand it by adding $P_2$ after we find that $A_8$ cannot be assigned to $P_4$ because of the choice of $A_6$ to support $P_2$. Informally, the conflict set representation can be seen as an incrementally maintained (partial) explanation of failure, indicating that there is a conflict between the current value of $P_2$ and one of the possible values of $P_4$ (Kambhampati, 1998).

We now consider the second possible value of $P_4$, viz., $A_9$, and find that it is mutually exclusive with $A_5$ which is currently supporting $P_1$. Following our practice, we add $P_1$ to the conflict set of $P_4$. At this point, there are no further choices for $P_4$, and so we backtrack from $P_4$, passing the conflict set of $P_4$, viz., $\{P_1, P_2, P_4\}$ as the reason for its failure. In essence, the conflict set is a shorthand notation for the following complete failure explanation (Kambhampati, 1998):[6]

$$[(P_4 = A_8) \lor (P_4 = A_9)] \land (P_1 = A_5 \Rightarrow P_4 \neq A_9) \land (P_2 = A_6 \Rightarrow P_4 \neq A_8) \land P_1 = A_5 \land P_2 = A_6$$

It is worth noting at this point that when $P_4$ is revisited in the future with different assignments to the preceding variables, its conflict set will be re-initialized to $\{P_4\}$ before considering any assignments to it.

The first advantage of the conflict set is that it allows a transparent way of supporting dependency directed backtracking (Kambhampati, 1998). In the current example, having failed to assign $P_4$, we have to start backtracking. We do not need to do this in a chronological fashion however.

---

6. We strip the first (disjunctive) clause since it is present in the graph structure, and the next two implicative clauses since they are part of the mutual exclusion relations that will not change for this problem. The conflict set representation just keeps the condition (variable) names of the last two clauses – denoting, in essence, that it is the current assignments of the variables $P_1$ and $P_2$ that are causing the failure to assign $P_4$.

**Find-Plan**($G$:**goals**, $pg$: **plan graph**, $k$: **level**)

    If $k = 0$, Return an empty subplan $P$ with success.

    If there is a memo $M$ such that $M \subseteq G$,

        Fail, and return $M$ as the conflict set

    Call *Assign-goals*$(G, pg, k, \emptyset)$.

        If Assign-goals fails and returns a conflict set $M$,

            Store $M$ as a memo

            *Regress* $M$ over actions selected at level $k + 1$ to get $R$

            Fail and return $R$ as the conflict set

        If Assign-goals succeeds, and returns a $k$-level subplan $P$,

            Return $P$ with success

**Assign-goals**($G$:**goals**, $pg$: **plan graph**, $k$: **level**, $A$: **actions**)

    If $G = \emptyset$

        Let $U$ be the union of preconditions of the actions in $A$

        Call *Find-plan*$(U, pg, k - 1)$

            If Find-plan fails and returns a conflict set $R$,

                Fail and return $R$

            If Find-plan succeeds and returns a subplan $P$ of length $k - 1$

                Succeed and return a $k$ length subplan $P \bullet A$

    Else ;;($G \neq \emptyset$)

        Select a goal $g \in G$

        Let $cs \leftarrow \{g\}$, and $A_g$ be the set of actions from level $k$ in $pg$ that support $g$

*L1:*        If $A_g = \emptyset$, Fail and return $cs$ as the conflict set

        Else, pick an action $a \in A_g$, and set $A_g \leftarrow A_g - a$

            If $a$ is mutually exclusive with some action $b \in A$

                Let $l$ be the goal that $b$ was selected to support

                Set $cs \leftarrow cs \cup \{l\}$

                Goto *L1*

            Else ($a$ is not mutually exclusive with any action in $A$)

                Call *Assign-goals*$(G - \{g\}, pg, k, A \cup \{a\})$

                    If the call fails and returns a conflict set $C$

                        If $g \in C$

                            Set $cs = cs \cup C$ *;conflict set absorption*

                            Goto *L1*

                        Else ;($g \notin C$)

                            Fail and return $C$ as the conflict set

                                  *;dependency directed backjumping*

Figure 4: A pseudo-code description of Graphplan backward search enhanced with EBL/DDB capabilities. The backward search at level $k$ in a planning-graph $pg$ is initiated with the call $Find\text{-}Plan(G, pg, k)$, where $G$ is the set of top level goals of the problem.

Instead, we jump back to the most recent variable (condition) taking part in the conflict set of $P_4$ – in this case $P_2$. By doing so, we are avoiding considering other alternatives at $P_3$, and thus avoiding one of the inefficiencies of the standard backward search. It is easy to see that such backjumping is sound since $P_3$ is not causing the failure at $P_4$ and thus re-assigning it won't avert the failure.

Continuing along, whenever the search backtracks to a condition $c$, the backtrack conflict is absorbed into the current conflict set of $c$. In our example, we absorb $\{P_1, P_2, P_4\}$ into the conflict set of $P_2$, which is currently $\{P_2\}$ (making $\{P_1, P_2, P_4\}$ the new conflict set of $P_2$). We now assign $A_{11}$, the only remaining value, to $P_2$. Next we try to assign $P_3$ and find that its only value $A_7$ is mutex with $A_{11}$. Thus, we set conflict set of $P_3$ to be $\{P_3, P_2\}$ and backtrack with this conflict set. When the backtracking reaches $P_2$, this conflict set is absorbed into the current conflict set of $P_2$ (as described earlier), giving rise to $\{P_1, P_2, P_3, P_4\}$ as the current combined failure reason for $P_2$. This step illustrates how the conflict set of a condition is incrementally expanded to collect the reasons for failure of the various possible values of the condition.

At this point, $P_2$ has no further choices, so we backtrack over $P_2$ with its current conflict set, $\{P_1, P_2, P_3, P_4\}$. At $P_1$, we first absorb the conflict set $\{P_1, P_2, P_3, P_4\}$ into $P_1$'s current conflict set, and then re-initiate backtracking since $P_1$ has no further choices.

Now, we have reached the end of the current level $(k - 1)$. Any backtracking over $P_1$ must involve undoing assignments of the conditions at the $k^{th}$ level. Before we do that however, we do two steps: memoization and regression.

## 4.1 Memoization

Before we backtrack over the first assigned variable at a given level, we store the conflict set of that variable as a memo at that level. We store the conflict set $\{P_1, P_2, P_3, P_4\}$ of $P_1$ as a memo at this level. Notice that the memo we store is shorter (and thus more general) than the one stored by the normal Graphplan, as we do not include $P_5$ and $P_6$, which did not have anything to do with the failure[7]

## 4.2 Regression

Before we backtrack out of level $k - 1$ to level $k$, we need to convert the conflict set of (the first assigned variable in) level $k - 1$ so that it refers to the conditions in level $k$. This conversion process involves regressing the conflict set over the actions selected at the $k^{th}$ level (Kambhampati, 1998). In essence, the regression step computes the (smallest) set of conditions (variables) at the $k^{th}$ level whose supporting actions spawned (activated, in DCSP terms) the conditions (variables) in the conflict set at level $k - 1$. In the current case, our conflict set is $\{P_1, P_2, P_3, P_4\}$. We can see that $P_2, P_3$ are required because of the condition $G_1$ at level $k$, and the condition $P_4$ is required because of the condition $G_2$.

In the case of condition $P_1$, both $G_1$ and $G_4$ are responsible for it, as both their supporting actions needed $P_1$. In such cases we have two heuristics for computing the regression: (1) Prefer choices that help the conflict set to regress to a smaller set of conditions (2) If we still have a choice between multiple conditions at level $k$, we pick the one that has been assigned earlier. The motivation for the first rule is to keep the failure explanations as compact (and thus as general) as possible, and the motivation for the second rule is to support deeper dependency directed backtracking. It is important to note that these heuristics are aimed at improving the performance of the EBL/DDB and do not affect the soundness and completeness of the approach.

---

7. While in the current example, the memo includes all the conditions up to $P_4$ (which is the farthest we have gone in this level), even this is not always necessary. We can verify that $P_3$ would not have been in the memo set if $A_{11}$ were not one of the supporters of $P_2$.

In the current example, the first of these rules applies, since $P_1$ is already required by $G_1$, which is also requiring $P_2$ and $P_3$. Even if this was not the case (i.e., $G_1$ only required $P_1$), we still would have selected $G_1$ over $G_4$ as the regression of $P_1$, since $G_1$ was assigned earlier in the search.

The result of regressing $\{P_1, P_2, P_3, P_4\}$ over the actions at $k^{th}$ level is thus $\{G_1, G_2\}$. We start backtracking at level $k$ with this as the conflict set. We jump back to $G_2$ right away, since it is the most recent variable named in the conflict set. This avoids the inefficiency of re-considering the choices at $G_3$ and $G_4$, as done by the normal backward search. At $G_2$, the backtrack conflict set is absorbed, and the backtracking continues since there are no other choices. Same procedure is repeated at $G_1$. At this point, we are once again at the end of a level–and we memoize $\{G_1, G_2\}$ as the memo at level $k$. Since there are no other levels to backtrack to, Graphplan is called on to extend the planning-graph by one more level.

Notice that the memos based on EBL analysis capture failures that may require a significant amount of search to rediscover. In our example, we are able to discover that $\{G_1, G_2\}$ is a failing goal set despite the fact that there are no mutex relations between the choices of the goals $G_1$ and $G_2$.

### 4.3 Using the Memos

Before we end this section, there are a couple of observations regarding the use of the stored memos. In the standard Graphplan, memos at each level are stored in a level-specific hash table. Whenever backward search reaches a level $k$ with a set of conditions to be satisfied, it consults the hash table to see if this exact set of conditions is stored as a memo. Search is terminated only if an exact hit occurs. Since EBL analysis allows us to store compact memos, it is not likely that a complete goal set at some level $k$ is going to exactly match a stored memo. What is more likely is that a stored memo is a subset of the goal set at level $k$ (which is sufficient to declare that goal set a failure). In other words, the memo checking routine in Graphplan needs to be modified so that it checks to see if some subset of the current goal set is stored as a memo. The naive way of doing it – which involves enumerating all the subsets of the current goal set and checking if any of them are in the hash table, turns out to be very costly. One needs more efficient data structures, such as the set-enumeration trees (Rymon, 1992). Indeed, Koehler and her co-workers (Koehler, Nebel, Hoffman, & Dimopoulos, 1997) have developed a data structure called UB-Trees for storing the memos. The UB-Tree structures can be seen as a specialized version of the "set-enumeration trees," and they can efficiently check if any subset of the current goal set has been stored as a memo.

The second observation regarding memos is that they can often serve as a failure explanation in themselves. Suppose we are at some level $k$, and find that the goal set at this level subsumes some stored memo $M$. We can then use $M$ as the failure explanation for this level, and regress it back to the previous level. Such a process can provide us with valuable opportunities for further back jumping at levels above $k$. It also allows us to learn new compact memos at those levels. Note that none of this would have been possible with normal memos stored by Graphplan, as the only way a memo can declare a goal set at level $k$ as failing is if the memo is exactly equal to the goal set. In such a case regression will just get us all the goals at level $k + 1$, and does not buy us any backjumping or learning power (Kambhampati, 1998).

## 5. Empirical Evaluation of the Effectiveness of EBL/DDB

We have now seen the way EBL and DDB capabilities are added to the backward search by maintaining and updating conflict-sets. We also noted that EBL and DDB capabilities avoid a variety of inefficiencies in the standard Graphplan backward search. That these augmentations are soundness and completeness preserving follows from the corresponding properties of conflict-directed backjumping (Kambhampati, 1998). The remaining (million-dollar) question is whether these capabilities make a difference in practice. I now present a set of empirical results to answer this question.

I implemented the EBL/DDB approach described in the previous section on top of a Graphplan implementation in Lisp.[8] The changes needed to the code to add EBL/DDB capability were relatively minor – only two functions needed non-trivial changes[9]. I also added the UB-Tree subset memo checking code described in (Koehler et al., 1997). I then ran several comparative experiments on the "benchmark" problems from (Kautz & Selman, 1996), as well as from four other domains. The specific domains included blocks world, rocket world, logistics domain, gripper domain, ferry domain, traveling salesperson domain, and towers of hanoi. Some of these domains, including the blocks world, the logistics domain and the gripper domain were used in the recent AI Planning Systems competition. The specifications of the problems as well as domains are publicly available.

Table 1 shows the statistics on the times taken and number of backtracks made by normal Graphplan, and Graphplan with EBL/DDB capabilities.[10]

### 5.1 Run-Time Improvement

The first thing we note is that EBL/DDB techniques can offer quite dramatic speedups – from 1.6x in blocks world all the way to 120x in the logistics domain (the Att-log-a problem is unsolvable by normal Graphplan after over 40 hours of cpu time!). We also note that the number of backtracks reduces significantly and consistently with EBL/DDB. Given the lengh of some of the runs, the time Lisp spends doing garbage collection becomes an important issue. I thus report the cumulative time (including cpu time and garbage collection time) for Graphplan with EBL/DDB, while I separate the cpu time from cumulative time for the plain Graphplan (in cases where the total time spent was large enough that garbage collection time is a significant fraction). Specifically, there are two entrys in the column corresponding to total time for the normal Graphplan. The first entry is the cpu time spent, while the second entry in parenthesis is the cumulative time (cpu time and garbage collection time) spent. The speedup is computed with respect to the cumulative time of Graphplan with EBL/DDB and cpu time of plain Graphplan. [11] The reported speedups should thus be seen as *conservative* estimates.

---

8. The original lisp implementation of Graphplan was done by Mark Peot. The implementation was subsequently improved by David Smith.

9. `Assign-goals` and `find-plan`

10. In the earlier versions of this paper, including the paper presented at IJCAI (Kambhampati, 1999) I have reported experiments on a Sun SPARC Ultra 1 running Allegro Common Lisp 4.3. The Linux machine run-time statistics seem to be approximately 2.7x faster than those from the Sparc machine.

11. It is interesting to note that the percentage of time spent doing garbage collection is highly problem dependent. For example, in the case of Att-log-a, only 30 minutes out of 41 hours (or about 1% of the cumulative time) was spent doing garbage collection, while in the case of Tower-6, 3.1 hours out of 4.8 hours (or about 65% of the cumulative time) was spent on garbage collection!

| Problem | Graphplan with EBL/DDB | | | | | Normal Graphplan | | | | | Speedup |
|---|---|---|---|---|---|---|---|---|---|---|---|
| | Tt | Mt | # Btks | AvLn | AvFM | Tt. | Mt. | # Btks | AvLn | AvFM | |
| Huge-Fact (18/18) | 3.08 | 0.28 | 2004K | *9.52* | 2.52 | 5.3 | 0.22 | 5181K | *11.3* | 1.26 | **1.7x** |
| BW-Large-B (18/18) | 2.27 | 0.11 | 798K | *10.15* | 3.32 | 4.15 | 0.05 | 2823K | *11.83* | 1.13 | **1.8x** |
| Rocket-ext-a (7/36) | .8 | .34 | 764K | *8.5* | 82 | 19.43 | 11.7 | 8128K | *23.9* | 3.2 | **24x** |
| Rocket-ext-b (7/36) | .8 | .43 | 569K | *7.5* | 101 | 14.1 | 7.7 | 10434K | *23.8* | 3.22 | **17x** |
| Att-log-a(11/79) | 1.97 | .89 | 2186K | *8.21* | 46.18 | >40.5hr (*>41hr*) | - | - | *32* | - | **>1215x** |
| Gripper-6 (11/17) | 0.1 | 0.03 | 201K | *6.9* | 6.2 | 1.1 | .39 | 2802K | *14.9* | 4.9 | **11x** |
| Gripper-8 (15/23) | 2.4 | .93 | 4426K | *9* | 7.64 | 215(*272*) | - | - | *17.8* | - | **90x** |
| Gripper-10(19/29) | 47.9 | 18.2 | 61373K | 11.05 | 8.3 | >8.2hr(*>16hr*) | - | - | - | - | **>10x** |
| Tower-5 (31/31) | .17 | 0.02 | 277K | *6.7* | 2.7 | 7.23 | 1.27 | 19070K | *20.9* | 2.2 | **42x** |
| Tower-6 (63/63) | 2.53 | 0.22 | 4098K | *7.9* | 2.8 | >1.7hr (*>4.8hr*) | - | - | *22.3* | - | **>40x** |
| Ferry-41 (27/27) | .44 | 0.13 | 723K | *7.9* | 2.54 | 22(*29*) | 11 | 33357K | *24.5* | 2.3 | **50x** |
| Ferry-5 (31/31) | 1.13 | .41 | 1993K | *8.8* | 2.53 | 42(*144*) | 24 | 53233K | *25* | 2.4 | **37x** |
| Ferry-6(39/39) | 11.62 | 5.3 | 18318K | 10.9 | 2.6 | >5hr(*>18.4hr*) | - | - | - | - | **>25x** |
| Tsp-10 (10/10) | .99 | 0.23 | 2232K | *6.9* | 12 | 89(*93*) | 56.7 | 68648K | *13* | 5 | **90x** |
| Tsp-12(12/12) | 12.4 | 2.65 | 21482K | 7.9 | 15.2 | >12hr (*>14.5hr*) | - | - | - | - | **>58x** |

Table 1: Empirical performance of EBL/DDB. Unless otherwise noted, times are in cpu minutes on a Pentium-III 500 MHZ machine with 256meg RA running Linux and allegro common lisp 5, compiled for speed. "Tt" is total time, "Mt" is the time used in checking memos and "Btks" is the number of backtracks done during search. The times for Graphplan with EBL/DDB include both the cpu and garbage collection time, while the cpu time is separated from the total time in the case of normal Graphplan. The numbers in parentheses next to the problem names list the number of time steps and number of actions respectively in the solution. AvLn and AvFM denote the average memo length and average number of failures detected per stored memo respectively.

## 5.2 Reduction in Memo Length

The results also highlight the fact that the speedups offered by EBL/DDB are problem/domain dependent – they are quite meager in blocks world problems, and are quite dramatic in many other domains including the rocket world, logistics, ferry, gripper, TSP and Hanoi domains. The statistics on the memos, shown in Table 1 shed light on the reasons for this variation. Of particular interest is the average length of the stored memos (given in the columns labeled "AvLn"). In general, we expect that the EBL analysis reduces the length of stored memos, as conditions that are not part of the failure explanation are not stored in the memo. However, the advantage of this depends on the likelihood that only a small subset of the goals at a given level are actually taking part in the failure. This likelihood in turn depends on the amount of inter-dependencies between the goals at a given level. From the table, we note that the average length reduces quite dramatically in the rocket world and logistics[12], while the reduction is much less pronounced in the blocks world. This variation can be traced back to a larger degree of inter-dependency between goals at a given level in the blocks world problems.

The reduction in average memo length is correlated perfectly with the speedups offered by EBL on the corresponding problems. Let me put this in perspective. The fact that the average length of memos for Rocket-ext-a problem is 8.5 with EBL and 24 without EBL, shows in essence that normal Graphplan is re-discovering an 8-sized failure embedded in $\binom{24}{8}$ possible ways in the worst case in a 24 sized goal set – storing a new memo each time (incurring both increased backtracking and matching costs)! It is thus no wonder that normal Graphplan performs badly compared to Graphplan with EBL/DDB.

## 5.3 Utility of Stored Memos

The statistics in Table 1 also show the increased utility of the memos stored by Graphplan with EBL/DDB. Since EBL/DDB store more general (smaller) memos than normal Graphplan, they should, in theory, generate fewer memos and use them more often. The columns labeled "AvFM" give the ratio of the number of failures discovered through the use of memos to the number of memos generated in the first place. This can be seen as a measure of the average "utility" of the stored memos. We note that the utility is consistently higher with EBL/DDB. As an example, in Rocket-ext-b, we see that on the average an EBL/DDB generated memo was used to discover failures 101 times, while the number was only 3.2 for the memos generated by the normal Graphplan.[13]

## 5.4 Relative Utility of EBL vs. DDB

From the statistics in Table 1, we see that even though EBL can make significant improvements in run-time, a significant fraction of the run time with EBL (as well as normal Graphplan) is spent in memo checking. This raises the possibility that the overall savings are mostly from the DDB part and that the EBL part (i.e, the part involving storing and checking memos) is in fact a net drain (Kambhampati, Katukam, & Qu, 1997). To see if this is true, I ran some problems with EBL (i.e., memo-checking) disabled. The DDB capability as well as the standard Graphplan memoization strategies were left in.[14] The results are shown in Table 2, and demonstrate that the ability to store smaller memos (as afforded by EBL) is quite helpful–giving rise to 120x speedup over DDB alone in the Att-log-a problem, and 50x speedup in Tower-6 problem. Of course, the results also show that DDB is an important capability in itself. Indeed, Att-log-aand tower-6 could not even be solved by the standard Graphplan, while with DDB, these problems become solvable. In summary, the results show that both EBL and DDB can have a net positive utility.

---

12. For the case of Att-log-a, I took the memo statistics by interrupting the search after about 6 hours

13. The statistics for Att-log-aseem to suggest that memo usage was not as bad for normal Graphplan. However, it should be noted that Att-log-a was not solved by normal Graphplan to begin with. The improved usage factor may be due mostly to the fact that the search went for a considerably longer time, giving Graphplan more opportunity to use its memos.

| Problem | EBL+DDB | | DDB | | Speedup |
|---|---|---|---|---|---|
| | Btks | Time | Btks | Time | |
| Att-log-a | 2186K | 1.95 | 115421K | 235 | 120x |
| Tower-6 | 4098K | 2.37 | 97395K | 121 | 51x |
| Rocket-ext-a | 764K | .83 | 3764K | 17.18 | 21x |
| Gripper-8 | 4426K | 2.43 | 5426K | 4.71 | 1.94x |
| TSP-10 | 2238K | 1.1 | 4308K | 2.3 | 2.09x |
| Huge-Fct | 2004K | 3.21 | 2465K | 3.83 | 1.19x |

Table 2: Utility of storing and using EBL memos over just doing DDB

### 5.5 Utility of Memoization

Another minor, but not well-recognized, point brought out by the statistics in Table 1 is that the memo checking can sometimes be a significant fraction of the run-time of standard Graphplan. For example, in the case of Rocket-ext-a, standard Graphplan takes 19.4 minutes of which 11.7 minutes, or over half the time, is spent in memo checking (in hash tables)! This raises the possibility that if we just disable the memoization, perhaps we can do just as well as the version with EBL/DDB. To see if this is the case, I ran some of the problems with memoization disabled. The results show that in general disabling memo-checking leads to worsened performance. While I came across some cases where the disablement reduces the overall run-time, the run-time is still much higher than what you get with EBL/DDB. As an example, in the case of Rocket-ext-a, if we disable the memo checking completely, Graphplan takes 16.5 minutes, which while lower than the 19.4 minutes taken by standard Graphplan, is still much higher than the .8 minutes taken by the version of Graphplan with EBL/DDB capabilities added. If we add DDB capability, while still disabling the memo-checking, the run time becomes 2.4 minutes, which is still 3 times higher than that afforded with EBL capability.

### 5.6 The C vs. Lisp Question

Given that most existing implementations of Graphplan are done in C with many optimizations, one nagging doubt is whether the dramatic speedups due to EBL/DDB are somehow dependent on the moderately optimized Lisp implementation I have used in my experiments. Thankfully, the EBL/DDB techniques described in this paper have also been (re)implemented by Maria Fox and Derek Long on their STAN system. STAN is a highly optimized implementation of Graphplan that fared well in the recent AIPS planning competition. They have found that EBL/DDB resulted in similar dramatic speedups on their system too (Fox, 1998; Fox & Long, 1999). For example, they were unable to solve Att-log-a with plain Graphplan, but could solve it easily with EBL/DDB added.

---

14. I also considered removing the memoization completely, but the results were even poorer.

Finally, it is worth pointing out that even with EBL/DDB capabilities, I was unable to solve some larger problems in the AT&T benchmarks, such as bw-large-c and Att-log-b. This is however not an indictment against EBL/DDB since to my knowledge the only planners that solved these problems have all used either local search strategies such as GSAT, randomized re-start strategies, or have used additional domain-specific knowledge and pre-processing. At the very least, I am not aware of any existing implementations of Graphplan that solve these problems.

## 6. On the Utility of Graphplan Memos

One important issue in using EBL is managing the costs of storage and matching. Indeed, as discussed in (Kambhampati, 1998), naive implementations of EBL/DDB are known to lose the gains made in pruning power in the matching and storage costs. Consequently, several techniques have been invented to reduce these costs through selective learning as well as selective forgetting. It is interesting to see why these costs have not been as prominent an issue for EBL/DDB on Graphplan. I think this is mostly because of two characteristics of Graphplan memoization strategy:

1. Graphplan's memoization strategy provides a very compact representation for no-goods, as well as a very selective strategy for remembering no-goods. Seen as DCSP, it only remembers subsets of activated variables that do not have a satisfying assignment. Seen as a CSP (c.f. Figure 2), Graphplan only remembers no-goods of the form

$$P_1^i \neq \perp \land P_2^i \neq \perp \cdots P_m^i \neq \perp$$

   (where the superscripts correspond to the level of the planning graph to which the proposition belongs), while normal EBL implementations learn no-goods of the form

$$P_1^i = a_1 \land P_2^j = a_2 \cdots P_m^k = a_m$$

   Suppose a planning graph contains $n$ propositions divided into $l$ levels, and each proposition $P$ at level $j$ has at most $d$ actions supporting it. A CSP compilation of the planning graph will have $n$ variables, each with $d+1$ values (the extra one for $\perp$). A normal EBL implementation for such a CSP can learn, in the worst case, $(d+2)^n$ no-goods.[15] In contrast, Graphplan remembers only $l * 2^{\frac{n}{l}}$ memos[16]–a very dramatic reduction. This reduction is a result of two factors:

   (a) Each individual memo stored by Graphplan corresponds to an exponentially large set of normal no-goods (the memo

$$P_1^i \neq \perp \land P_2^i \neq \perp \cdots P_m^i \neq \perp$$

   is a shorthand notation for the conjunction of $d^m$ no-goods corresponding to all possible non- $\perp$ assignments to $P_1^i \cdots P_m^i$)

---

15. Each variable $v$ may either not be present in a no-good, or be present with one of $d+1$ possible assignments–giving $d+1$ possibilities for each of $n$ variables.

16. At each level, each of $\frac{n}{l}$ propositions either occurs in a memo or does not occur

   (b) Memos only subsume no-goods made up of proposition variables from the same planning graph level.

2. The matching cost is reduced by both the fact that considerably fewer no-goods are ever learned, and the fact that Graphplan stores no-goods (memos) separately for each level, and only consults the memos stored at level $j$, while doing backwards search at a level $j$,

The above discussion throws some light on why the so-called "EBL utility" problem is not as critical for Graphplan as it is for EBL done on normal CSPs.

## 6.1 Scenarios Where Memoization is too Conservative to Avoid Rediscovery of the Same Failures

The discussion above also raises the possibility that Graphplan (even with EBL/DDB) memoization is too conservative and may be losing some useful learning opportunities only because they are not in the required syntactic form. Specifically, before Graphplan can learn a memo of the form

$$P_1^i \neq \perp \wedge P_2^i \neq \perp \cdots P_m^i \neq \perp,$$

it must be the case that *each* of the $d^m$ possible assignments to the $m$ propositional variables must be a no-good. Even if *one* of them is not a no-good, Graphplan avoids learning the memo, thus potentially repeating the failing searches at a later time (although the loss is made up to some extent by learning several memos at a lower level).

Consider for example the following scenario: we have a set of variables $P_1^i \cdots P_m^i \cdots P_n^i$ at some level $i$ that are being assigned by backward search. Suppose the search has found a legal partial assignment for the variables $P_1^i \cdots P_{m-1}^i$, and the domain of $P_m^i$ contains the $k$ values $\{v_1 \cdots v_k\}$. In trying to assign the variables $P_m^i \cdots P_n^i$, suppose we repeatedly fail and backtrack up to the variable $P_m^i$, re-assigning it and eventually settling at the value $v_7$. At this point once again backtracking occurs, but this time we backtrack *over $P_m^i$* to higher level variables $(P_1^i \cdots P_m^i)$ and re-assigning them. At this point, it would have been useful to remember some no-goods to the effect that none of the first 6 values of $P_m^i$ are going to work so all that backtracking does not have to be repeated. Such no-goods will take the form:

$$P_m^i = v_j \wedge P_{m+1}^i \neq \perp \wedge P_{m+2}^i \neq \perp \cdots P_n^i \neq \perp$$

where $j$ ranges over $1 \cdots 6$, for all the values of $P_m^i$ that were tried and found to lead to failure while assigning the later variables. Unfortunately, such no-goods are not in the syntactic form of memos and so the memoization procedure cannot remember them. The search is thus forced to rediscover the same failures.

## 6.2 Sticky Values as a Partial Antidote

One way of staying with the standard memoization, but avoiding rediscovery of the failing search paths, such as those in the case of the example above, is to use the "sticky values" heuristic (Frost & Dechter, 1994; Kambhampati, 1998). This involves remembering the current value of a variable while skipping over it during DDB, and trying that value first when the search comes back to that variable. The heuristic is motivated by the fact that when we skip over a variable during DDB, it means that the variable and its current assignment have not contributed to the failure that caused the backtracking–so it makes sense to restore this value upon re-visit. In the example above, this heuristic will remember that $v_7$ was the current value of $P_m^i$ when we backtracked over it, and tries that as the first value when it is re-visited. A variation on this technique is to re-arrange or *fold* the domain of the variable such that all the values that precede the current value are sent to the back of the domain, so that these values will be tried only if other previously untried values are found to fail. This makes the assumption that the values that led to failure once are likely to do so again. In the example above, this heuristic folds the domain of $P_m^i$ so it becomes $\{v_7, v_8 \cdots v_k, v_1, v_2 \cdots v_6\}$. Notice that both these heuristics make sense only if we employ DDB, as otherwise we will never skip over any variable during backtracking.

I implemented both sticky value heuristics on top of EBL/DDB for Graphplan. The statistics in Table 3 show the results of experiments with this extension. As can be seen, the sticky values approach is able to give up to 4.6x additional speedup over EBL/DDB depending on the problem. Further, while the folding heuristic dominates the simple version in terms of number of backtracks, the difference is quite small in terms of run-time.

## 7. Forward Checking & Dynamic Variable Ordering

DDB and EBL are considered "look-back" techniques in that they analyze the failures by looking back at the past variables that may have played a part in those failures. There is a different class of techniques known as "look-forward" techniques for improving search. Prominent among these latter are forward checking and dynamic variable ordering. Supporting forward checking involves filtering out the conflicting actions from the domains of the remaining goals, as soon as a particular goal is assigned. In the example in Figure 1, forward checking will filter $A_9$ from the domain of $P_4$ as soon as $P_1$ is assigned $A_5$. Dynamic variable ordering (DVO) involves selecting for assignment the goal that has the least number of remaining establishers.[17] When DVO is combined with forward checking, the variables are ordered according to their "live" domain sizes (where live domain is comprised of values from the domain that are not yet pruned by forward checking). Our experiments[18] show that these techniques can bring about reasonable, albeit non-dramatic, improvements in Graphplan's performance. Table 4 shows the statistics for some benchmark problems, with dynamic variable ordering alone, and with forward checking and dynamic variable ordering. We note that while the backtracks reduce by up to 3.6x in the case of dynamic variable ordering, and 5x in the case of dynamic variable ordering and forward checking, the speedups in time are somewhat smaller, ranging only from 1.1x to 4.8x. Times can perhaps be improved further with a more efficient implementation of forward checking.[19] The results also seem to suggest that no amount of optimization is going to make dynamic variable ordering and forward checking competitive with EBL/DDB on other problems. For one thing, there are several problems, including Att-log-a, Tsp-12, Ferry-6 etc. which just could not be solved even with forward checking and dynamic variable ordering. Second, even on the problems that could be solved, the reduction in backtracks provided by EBL/DDB is far greater than that provided by FC/DVO strategies. For example, on Tsp-10, the FC/DVO strategies reduce number of backtracks from 69974K to 14697K, a 4.8x improvement. However, this pales in comparison to 2232K backtracks (or 31x improvement) given by by EBL/DDB (see the entry in Table 1). Notice that these results only say that variable ordering strategies do not make a dramatic difference for Graphplan's backward search (or a DCSP compilation of the planning graph); they do not make any claims about the utility of FC and DVO for a CSP compilation of the planning graph.

---

17. I have also experimented with a variation of this heuristic, known as the Brelaz heuristic (Gomes et al., 1998), where the ties among variables with the same sized live-domains are broken by picking variables that take part in most number of constraints. This variation did not however lead to any appreciable improvement in performance.

18. The study of forward checking and dynamic variable ordering was initiated with Dan Weld.

19. My current implementation physically removes the pruned values of a variable during forward checking phase, and restores values on backtracks. There are better implementations, including use of in/out flags on the values as well as use of indexed arrays (c.f. (Bacchus & van Run, 1995))

| Problem | Plain EBL/DDB | | EBL/DDB+Sticky | | | EBL/DDB+Sticky+Fold | | |
|---|---|---|---|---|---|---|---|---|
| | Time | Btks | Time | Btks | Speedup | Time | Btks | Speedup |
| Rocket-ext-a(7/36) | .8 | 764K | .37 | 372K | 2.2x(2.05x) | .33 | 347K | 2.4x (2.2x) |
| Rocket-ext-b(7/36) | .8 | 569K | .18 | 172K | 4.6x(3.3x) | .177 | 169K | 4.5x(3.36x) |
| Gripper-10(39/39) | 47.95 | 61373K | 36.9 | 56212K | 1.29x(1.09x) | 40.8 | 54975K | 1.17x(1.12x) |
| Ferry-6 | 11.62 | 18318K | 11.75 | 18151K | .99x(1.01x) | 11.87 | 18151K | .97x(1.01x) |
| TSP-12(12/12) | 12.44 | 21482K | 9.86 | 20948K | 1.26x(1.02x) | 10.18 | 20948K | 1.22x(1.02x) |
| Att-log-a(11/79) | 1.95 | 2186K | .95 | 1144K | 2x(1.91x) | .67 | 781K | 2.9x(2.8x) |

Table 3: Utility of using sticky values along with EBL/DDB.

| Problem | GP | GP+DVO | Speedup | GP+DVO+FC | Speedup |
|---------|-----|--------|---------|-----------|---------|
| Huge-fact (18/18) | 5.3(5181K) | 2.26 (1411K) | 2.3x(3.6x) | 3.59 (1024K) | 1.47x(5x) |
| BW-Large-B (18/18) | 4.15(2823K) | 3.14(1416K) | 1.3x(2x) | 4.78(949K) | .86(3x) |
| Rocket-ext-a (7/36) | 19.43(8128K) | 14.9(5252K) | 1.3x(1.5x) | 14.5(1877K) | 1.3x(4.3x) |
| Rocket-ext-b (7/36) | 14.1(10434K) | 7.91(4382K) | 1.8x(2.4x) | 6(1490K) | 2.4x(7x) |
| Att-log-a(11/79) | >10hr | >10hr | - | >10hr. | |
| Gripper-6(11/17) | 1.1(2802K) | .65(1107K) | 1.7x(2.5x) | .73 (740K) | 1.5x(3.7x) |
| Tsp-10(10/10) | 89(69974K) | 78(37654K) | 1.14x(1.9x) | 81(14697K) | 1.09x(4.8x) |
| Tower-6(63/63) | >10hr | >10hr | - | >10hr. | |

Table 4: Impact of forward checking and dynamic variable ordering routines on Graphplan. Times are in cpu minutes as measured on a 500 MHZ Pentium-III running Linux and Franz Allegro Common Lisp 5. The numbers in parentheses next to times are the number of backtracks. The speedup columns report two factors–the first is the speedup in time, and the second is the speedup in terms of number of backtracks. While FC and DVO tend to reduce the number of backtracks, the reduction does not always seem to show up in the time savings.

## 7.1 Complementing EBL/DDB with Forward Checking and Dynamic Variable Ordering

Although forward checking and dynamic variable ordering approaches were not found to be particularly effective in isolation for Graphplan's backward search, I thought that it would be interesting to revisit them in the context of a Graphplan enhanced with EBL/DDB strategies. Part of the original reasoning underlying the expectation that goal (variable) ordering will not have a significant effect on Graphplan performance is based on the fact that all the failing goal sets are stored in-toto as memos (Blum & Furst, 1997, pp. 290). This reason no longer holds when we use EBL/DDB. Further more, there exists some difference of opinion as to whether or not forward checking and DDB can fruitfully co-exist. The results of (Prosser, 1993) suggest that domain-filtering–such as the one afforded by forward checking, degrades intelligent backtracking. The more recent work (Frost & Dechter, 1994; Bayardo & Schrag, 1997) however seems to suggest however that best CSP algorithms should have both capabilities.

While adding plain DVO capability on top of EBL/DDB presents no difficulties, adding forward checking does require some changes to the algorithm in Figure 4. The difficulty arises because a failure may have occurred as a combined effect of the forward checking and backtracking. For example, suppose we have four variables $v_1 \cdots v_4$ that are being considered for assignment in that order. Suppose $v_3$ has the domain $\{1, 2, 3\}$, and $v_3$ cannot be 1 if $v_1$ is $a$, and cannot be 2 if $v_2$ is $b$. Suppose further that $v_4$'s domain only contains $d$, and there is a constraint saying that $v_4$ can't

| Problem | EBL | EBL+DVO | | EBL+FC+DVO | |
|---------|-----|---------|---|-----------|---|
| | Time(btks) | Time(btks) | Speedup | Time(Btks) | Speedup |
| Huge-fct | 3.08(2004K) | 1.51(745K) | 2x(2.68x) | 2.57(404K) | 1.2x(5x) |
| BW-Large-B | 2.27(798K) | 1.81(514K) | 1.25x(1.6x) | 2.98(333K) | .76x(2.39x) |
| Rocket-ext-a | .8(764K) | .4(242K) | 2x(3.2x) | .73(273K) | 1.09x(2.8x) |
| Rocket-ext-b | .8(569K) | .29(151K) | 2.75x(3.76x) | .72(195K) | 1.1x(2.9x) |
| Att-log-a | 1.97(2186K) | 2.59(1109K) | .76x(1.97x) | 3.98(1134K) | .5x(1.92x) |
| Tower-6 | 2.53(4098K) | 3.78(3396K) | .67x(1.2x) | 2.09(636K) | 1.2x(6.4x) |
| TSP-10 | .99(2232K) | 1.27(1793K) | .77x(1.24x) | 1.34(828K) | .73x(2.7x) |

Table 5: Effect of complementing EBL/DDB with dynamic variable ordering and forward checking strategies. The speedup columns report two factors–the first is the speedup in time, and the second is the speedup in terms of number of backtracks. While FC and DVO tend to reduce the number of backtracks, the reduction does not always seem to show up in the time savings.

be $d$ if $v_1$ is $a$ and $v_3$ is 3. Suppose we are using forward checking, and have assigned $v_1, v_2$ the values $a$ and $b$. Forward checking prunes 1 and 2 from $v_3$'s domain, leaving only the value 3. At this point, we try to assign $v_4$ and fail. If we use the algorithm in Figure 4, the conflict set for $v_4$ would be $\{v_4, v_3, v_1\}$, as the constraint that is violated is $v_1 = a \land v_3 = 3 \land v_4 = d$. However this is not sufficient since the failure at $v_4$ may not have occurred if forward checking had not stripped the value 2 from the domain of $v_3$. This problem can be handled by pushing $v_1$ and $v_2$, the variables whose assignment stripped some values from $v_3$, into $v_3$'s conflict set.[20] Specifically, the conflict set of every variable $v$ is initialized to $\{v\}$ to begin with, and whenever $v$ loses a value during forward checking with respect to the assignment of $v'$, $v'$ is added to the conflict set of $v$. Whenever a future variable (such as $v_4$) conflicts with $v_3$, we add the conflict set of $v_3$ (rather than just $v_3$) to the conflict set of $v_4$. Specifically the line

"Set cs = cs $\cup$ { $l$ }"

in the procedure in Figure 4 is replaced with the line

"Set cs = cs $\cup$ Conflict-set($l$)"

I have incorporated the above changes into my implementation, so it can support support forward checking, dynamic variable ordering as well as EBL on Graphplan. Table 5 shows the performance of this version on the experimental test suite. As can be seen from the numbers, the number of backtracks are reduced by up to 3.7x in the case of EBL+DVO, and up to 5x in the case of EBL+FC+DVO. The cpu time improvements are somewhat lower. While we got up to 2.7x speedup

---

20. Notice that it is possible that the values that were stripped off from $v_3$'s domain may not have had any impact on the failure to assign $v_4$. For example, perhaps there is another constraint that says that $v_4$ can't be $d$ if $v_3$ is $b$, and in that case, strictly speaking, the assignment of $v_2$ cannot really be blamed for the failure at $v_4$. While this leads to non-minimal explanations, there is no reason to expect that strict minimization of explanations is a pre-requisite for the effectiveness of EBL/DDB; see (Kambhampati, 1998) with EBL+DVO, and up to 1.2x speedup with EBL+FC+DVO, in several cases, the cpu times *increase* with FC and DVO. Once again, I attribute this to the overheads of forward checking (and to a lesser extent, of dynamic variable ordering). Most importantly, by comparing the results in the Tables 4 and 5, we can see that EBL/DDB capabilities are able to bring about significant speedups even over a Graphplan implementation using FC and DVO.

## 8. EBL/DDB & Randomized Search

Recent years have seen increased use of randomized search strategies in planning. These include both purely local search strategies (Gerevini, 1999; Selman, Levesque, & Mitchell, 1992) as well as hybrid strategies that introduce a random restart scheme on top of a systematic search strategy (Gomes et al., 1998). The BLACKBOX planning system (Kautz & Selman, 1999) supports a variety of random restart strategies on top of a SAT compilation of the planning graph, and empirical studies show that these strategies can, probabilistically speaking, scale up much better than purely systematic search strategies.

I wanted to investigate if (and by how much) EBL & DDB techniques will help Graphplan even in the presence of these newer search strategies. While EBL and DDB techniques have little applicability to purely local search strategies, they could in theory help random restart systematic search strategies. Random restart strategies are motivated by an attempt to exploit the "*heavy-tail*" distribution (Gomes et al., 1998) of the solution nodes in the search trees of many problems. Intuitively, in problems where there are a non-trivial percentage of very easy to find solutions as well as very hard to find solutions, it makes sense to restart the search when we find that we are spending too much effort for a solution. By restarting this way, we hope to (probabilistically) hit on the easier-to-find solutions.

I implemented a random-restart strategy on top of Graphplan by making the following simple modifications to the backward search:

1. We keep track of the number of times the backward search backtracks from one level of the plan graph to a previous level (a level closer to the goal state), and whenever this number exceeds a given limit (called *backtrack limit*), the search is restarted (by going back to the last level of the plan graph), assuming that the number of restarts has not also exceeded the given limit. The search process between any two restarts is referred to as an *epoch*.

2. The supporting actions (values) for a proposition variable are considered in a randomized order. It is this randomization that ensures that when the search is restarted, we will look at the values of each variable in a different order.[21]

Notice that random-restart strategy still allows the application of EBL and DDB strategies, since during any given epoch, the behavior of the search is identical to that of the standard backward search algorithm. Indeed, as the backtrack limit and the number of restarts are made larger and larger, the whole search becomes identical to standard backward search.

---

21. Reordering values of a variable doesn't make a whole lot of sense in BLACKBOX which is based on SAT encodings and thus has only boolean variables. Thus, the randomization in BLACKBOX is done on the order in which the goals are considered for assignment. This typically tends to clash with the built-in goal ordering strategies (such as DVO and SAT-Z (Li & Anbulagan, 1997)), and they get around this conflict by breaking ties among variables randomly. To avoid such clashes, I decided to randomize Graphplan by reordering values of a variable. I also picked inter-level backtracks as a more natural parameter characterizing the difficulty of a problem for Graphplan's backward search.

| Problem | Parameters R/B/L | Graphplan with EBL/DDB | | | | Normal Graphplan | | | |
|---------|------------------|------|--------|------|---------|------|--------|------|---------|
| | | %sol | Length | Time | Av. MFSL | %sol | Length | Time | Av. MFSL |
| Att-log-a(11/54) | 5/50/20 | 99% | 14(82) | .41 | 4.6K(28K) | 2% | 19(103) | .21 | .3K(3.7K) |
| Att-log-a(11/54) | 10/100/20 | 100% | 11.3(69.5) | .72 | 17.8K(59K) | 11% | 17.6(100.5) | 1.29 | 3.7K(41K) |
| Att-log-a(11/54) | 10/100/30 | 100% | 11.3(69.5) | .72 | 17.8K(59K) | 54% | 25.6(136) | 3 | 4K(78K) |
| Att-log-a(11/54) | 20/200/20 | 100% | 11(68.5) | 2.38 | 73K(220K) | 13% | 18(97.5) | 3 | 31K(361K) |
| Att-log-a(11/54) | 20/200/30 | 100% | 11(68.5) | 2.38 | 73K(220K) | 94% | 22.1(119.3) | 31 | 33K(489K) |
| Att-log-b(13/47) | 5/50/20 | 17% | 18.1(101) | 1.62 | 8K(93K) | 0% | - | - | .2K(4K) |
| Att-log-b(13/47) | 10/100/20 | 60% | 17.3(98) | 11.4 | 69K(717K) | 0% | - | - | 2.6K(53K) |
| Att-log-b(13/47) | 10/100/30 | 100% | 20.1(109) | 15.3 | 74K(896K) | 3% | 28(156) | 4 | 5K(111K) |
| Att-log-c(13/65) | 5/50/30 | 55% | 22.85(124) | 2.77 | 8K(145K) | 2% | 26.5(135) | .75 | .4K(8K) |
| Att-log-c(13/65) | 10/100/30 | 100% | 19.9(110) | 14 | 71K(848K) | 2% | 29(152) | 4 | 3.7K(111K) |
| Rocket-ext-a(7/34) | 10/100/30 | 100% | 7.76(35.8) | 1.3 | 29K(109K) | 58% | 21.24(87.3) | 2 | .2K(4K) |
| Rocket-ext-a(7/34) | 20/200/30 | 100% | 7(34.1) | 1.32 | 38K(115K) | 90% | 21.3(85) | 8.1 | 2.3K(43K) |
| Rocket-ext-a(7/34) | 40/400/30 | 100% | 7(34.2) | 1.21 | 35K(105K) | 100% | 15.3(62.5) | 45 | 35K(403K) |

Table 6: Effect of EBL/DDB on random-restart Graphplan. Time is measured in cpu minutes on Allegro Common Lisp 5.0 running on a Linux 500MHZ Pentium machine. The numbers next to the problem names are the number of steps and actions in the shortest plans reported for those problems in the literature. The R/B/L parameters in the second column refer to the limits on the number of restarts, number of backtracks and the number levels to which the plan graph is expanded. All the statistics are averaged over multiple runs (typically 100 or 50). The "MFSL" column gives the average number of memo-based failures per searched level of the plan graph. The numbers in parentheses are the total number of memo-based failures averaged over all runs. Plan lengths were averaged only over the successful runs.

To check if my intuitions about the effectiveness of EBL/DDB in randomized search were indeed correct, I conducted an empirical investigation comparing the performance of random search on standard Graphplan as well as Graphplan with EBL/DDB capabilities. Since the search is randomized, each problem is solved multiple number of times (100 times in most cases), and the runtime, plan length and other statistics were averaged over all the runs. The experiments are conducted with a given backtrack limit, a given restart limit, as well as a limit on the number of levels to which the planning graph is extended. This last one is needed as in randomized search, a solution may be missed at the first level it appears, leading to a prolonged extension of the planning graph until a (inoptimal) solution is found at a later level. When the limit on the number of levels is expanded, the probability of finding solution increases, but at the same time, the cpu time spent searching the graph also increases.

Having implemented this random restart search, the first thing I noticed is an improvement in the solvability horizon (as expected, given the results in (Gomes et al., 1998)). Table 6 shows these results. One important point to note is that the results in the table above talk about *average* plan lengths and cpu times. This is needed as due to randomization potentially each run can produce a different outcome (plan). Secondly, while Graphplan with systematic search guarantees shortest plans (measured in the number of steps), the randomized search will not have such a guarantee. In particular, the randomized version might consider a particular planning graph to be barren of solutions, based simply on the fact that no solution could be found within the confines of the given backtrack limit and number of restarts.

Graphplan, with or without EBL/DDB, is more likely to solve larger problems with randomized search strategies. For example, in the logistics domain, only the Att-log-a problem was solvable (within 24 hours real time) with EBL and systematic search. With the randomization added, my implementation was able to solve both Att-log-b and Att-log-c quite frequently. As the limits on the number of restarts, backtracks and levels is increased, the likelihood of finding a solution as well as the average length of the solution found improves. For example, Graphplan with EBL/DDB is able to solve Att-log-b in every trial for 10 restarts, 100 backtracks and 30 levels as the limits (although the plans are quite inoptimal).

The next, and perhaps more interesting, question I wanted to investigate is whether EBL and DDB will continue to be useful for Graphplan when it uses randomized search. At first blush, it seems as if they will not be as important–after all even Graphplan with standard search may luck out and be able to find solutions quickly in the presence of randomization. Further thought however suggests that EBL and DDB may still be able to help Graphplan. Specifically, they can help Graphplan in using the given backtrack limit in a more judicious fashion. To elaborate, suppose the random restart search is being conducted with 100 backtracks and 10 restarts. With EBL and DDB, Graphplan is able to pinpoint the cause of the failure more accurately than without EBL and DDB. This means that when the search backtracks, the chance that it will have to backtrack again for the same (or similar) reasons is reduced. This in turn gives the search more of a chance on catching a success during one of the number of epochs allowed. All this is in addition to the more direct benefit of being able to use the stored memos across epochs to cut down search.

As can be seen from the data in Table 6, for a given set of limits on number of restarts, number of backtracks, and number of levels expanded, Graphplan with EBL/DDB is able to get a higher percentage of solvability as well as significantly shorter length solutions (both in terms of levels and in terms of actions). To get comparable results on the standard Graphplan, I had to significantly increase the input parameters (restarts, backtracks and levels expanded), which in turn led to dra-matic increases in the average run time. For example, for the Att-log-a problem, with 5 restarts and 50 backtracks, and 20 levels limit, Graphplan was able to solve the problem 99% of the time, with an average plan length of 14 steps and 82 actions. In contrast, without EBL/DDB, Graphplan was able to solve the problem in only 2% of the cases, with an average plan length of 19 steps and 103 actions. If we double the restarts and backtracks, the EBL/DDB version goes to 100% solvability with an average plan length of 11.33 steps and 69.53 actions. The standard Graphplan goes to 11% solvability and a plan length of 17.6 steps and 100 actions. If we increase the number of levels to 30, then the standard Graphplan solves 54% of the problems with an average plan length of 25.6 steps and 136 actions. It takes 20 restarts and 200 backtracks, as well as a 30-level limit before standard Graphplan is able to cross 90% solvability. By this time, the average run time is 31 minutes, and the average plan length is 22 steps and 119 actions. The contrast between this and the 99% solvability in 0.4 minutes with 14 step 82 action plans provided by Graphplan with EBL and 5 restarts and 50 backtracks is significant! Similar results were observed in other problems, both in logistics (Att-log-b, Att-log-c) and other domains (Rocket-ext-a, Rocket-ext-b).

The results also show that Graphplan with EBL/DDB is able to generate and reuse memos effectively across different restart epochs. Specifically, the numbers in the columns titled "Av. MFSL" give the average number of *m*emo-based *f*ailures per *s*earch *l*evel.[22] We note that in all cases, the average number of memo-based failures are significantly higher for Graphplan with EBL than for normal Graphplan. This shows that EBL/DDB analysis is helping Graphplan reduce wasted effort significantly, and thus reap better benefits out of the given backtrack and restart limits.

## 9. Related Work

In their original implementation of Graphplan, Blum and Furst experimented with a variation of the memoization strategy called "subset memoization". In this strategy, they keep the memo generation techniques the same, but change the way memos are used, declaring a failure when a stored memo is found to be a subset of the current goal set. Since complete subset checking is costly, they experimented with a "partial" subset memoization where only subsets of length $n$ and $n-1$ are considered for an $n$ sized goal set.

As we mentioned earlier, Koehler and her co-workers (Koehler et al., 1997) have re-visited the subset memoization strategy, and developed a more effective solution to complete subset checking that involves storing the memos in a data structure called UB-Tree, instead of in hash tables. The results from their experiments with subset memoization are mixed, indicating that subset memoization does not seem to improve the cpu time performance significantly. The reason for this is quite easy to understand – while they improved the memo checking time with the UB-Tree data structure, they are still generating and storing the same old long memos. In contrast, the EBL/DDB extension described here supports dependency directed backtracking, and by reducing the average length of stored memos, increases their utility significantly, thus offering dramatic speedups.

To verify that the main source of power in the EBL/DDB-Graphplan is in the EBL/DDB part and not in the UB-Tree based memo checking, I re-ran my experiments with EBL/DDB turned off, but with subset memo checking with UB-Tree data structure still enabled. The results are shown in in Table 7. The columns labeled "AvFM" show that as expected subset memoization does improve the utility of stored memos over normal Graphplan (since it uses a memo in more scenarios than normal Graphplan can). However, we also note that subset memoization by itself does not have any dramatic impact on the performance of Graphplan, and that EBL/DDB capability can significantly enhance the savings offered by subset memoization.

---

22. Notice that the number of search levels may be different from (and smaller than) the number of planning graph levels, because Graphplan initiates a search only when none of the goals are pair-wise mutex with each other. In Att-log-a, Att-log-b and Att-log-c, this happens starting at level 9. For Rocket-ext-a it happens starting at level 5. The numbers in parentheses are the total number of memo based failures. We divide this number by the average number of levels in which search was conducted to get the "Av. MFSL" statistic.

| Problem | Tt | Mt | #Btks | **EBL x↑** | #Gen | #Fail | AvFM | AvLn |
|---------|-----|------|-------|---------|--------|--------|------|-------|
| Huge-Fact | 3.20 | 1 | 2497K | **1.04x** | 24243 | 33628 | 1.38 | 11.07 |
| BW-Large-b | 2.74 | 0.18 | 1309K | **1.21x** | 11708 | 15011 | 1.28 | 11.48 |
| Rocket-ext-a | 19.2 | 16.7 | 6188K | **24x** | 62419 | 269499 | 4.3 | 24.32 |
| Rocket-ext-b | 7.36 | 4.77 | 7546K | **9.2x** | 61666 | 265579 | 4.3 | 24.28 |
| Att-log-a | > 12hrs | - | - | **>120x** | - | - | - | - |

Table 7: Performance of subset memoization with UB-Tree data structure (without EBL/DDB). The "Tt" is the total cpu time and "Mt" is the time taken for checking memos. "#Btks" is the number of backtracks. "EBLx" is the amount of speedup offered by EBL/DDB over subset memoization "#Gen" lists the number of memos generated (and stored), "#Fail" lists the number of memo-based failures, "AvFM" is the average number of failures identified per generated memo and "AvLn" is the average length of stored memos.

In (Kambhampati, 1998), I describe the general principles underlying the EBL/DDB techniques and sketch how they can be extended to dynamic constraint satisfaction problems. The development in this paper can be seen as an application of the ideas there. Readers needing more background on EBL/DDB are thus encouraged to review that paper. Other related work includes previous attempts at applying EBL/DDB to planning algorithms, such as the work on UCPOP+EBL system (Kambhampati et al., 1997). One interesting contrast is the ease with which EBL/DDB can be added to Graphplan as compared to UCPOP system. Part of the difference comes from the fact that the search in Graphplan is ultimately on a propositional dynamic CSP, while in UCPOP's search is a variablized problem-solving search.

As I mentioned in Section 2, Graphplan planning graph can also be compiled into a normal CSP representation, rather than the dynamic CSP representation. I used the dynamic CSP representation as it corresponds quite directly to the backward search used by Graphplan. We saw that the model provides a clearer picture of the mutex propagation and memoization strategies, and helps us unearth some of the sources of strength in the Graphplan memoization strategy–including the fact that memos are a very conservative form of no-good learning that obviate the need for the no-good management strategies to a large extent.

The dynamic CSP model may also account for some of the peculiarities of the results of my empirical studies. For example, it is widely believed in the CSP literature that forward checking and dynamic variable ordering are either as critical as, or perhaps even more critical than, the EBL/DDB strategies (Bacchus & van Run, 1995; Frost & Dechter, 1994). Our results however show that for Graphplan, which uses the dynamic CSP model of search, DVO and FC are largely ineffective compared to EBL/DDB on the standard Graphplan. To some extent, this may be due to the fact that Graphplan already has a primitive form of EBL built into its memoization strategy. In fact, Blum & Furst (1997) argue that with memoization and a minimal action set selection (an action set is considered minimal if it is not possible to remove an action from the set and still support all the goals for which the actions were selected), the ordering of goals will have little effect (especially in the earlier levels that do not contain a solution).

Another reason for the ineffectiveness of the dynamic variable ordering heuristic may have to do with the differences between the CSP and DCSP problems. In DCSP, the main aim is not just to quickly find an assignment for the the current level variables, but rather to find an assignment for the current level which is likely to activate fewer and easier to assign variables, whose assignment in turn leads to fewer and easier to assign variables and so on. The general heuristic of picking the variable with the smallest (live) domain does not necessarily make sense in DCSP, since a variable with two actions supporting it may actually be much harder to handle than another with many actions supporting it, if each of the actions supporting the first one eventually lead to activation of many more and harder to assign new variables. It may thus be worth considering ordering strategies that are more customized to the dynamic CSP models–e.g. orderings that are based on the number (and difficulty) of variables that get activated by a given variable (or value) choice.

We have recently experimented with a value-ordering heuristic that picks the value to be assigned to a variable using the distance estimates of the variables that will be activated by that choice (Kambhampati & Nigenda, 2000). The planning graph provides a variety of ways of obtaining these distance estimates. The simplest idea would be to say that the distance of a proposition $p$ is the level at which $p$ enters the planning graph for the first time. This distance estimate can then be used to rank variables and their values. Variables can be ranked simply in terms of their distances–the variables that have the highest distance are chosen first (akin to fail-first principle). Value ordering is a bit trickier–for a given variable, we need to pick an action whose precondition set has the lowest distance. The distance of the precondition set can be computed from the distance of the individual preconditions in several ways:

- Maximum of the distances of the individual propositions making up the preconditions.

- Sum of the distances of the individual propositions making up the preconditions.

- The first level at which the set of propositions making up the preconditions are present and are non-mutex.

In (Kambhampati & Nigenda, 2000), we evaluate goal and value ordering strategies based on these ideas, and show that they can lead to quite impressive (upto 4 orders of magnitude in our tests) speedups in solution-bearing planning graphs. We also relate the distances computed through planning graph to the distance transforms computed by planners like HSP (Bonet, Loerincs, & Geffner, 1999) and UNPOP (McDermott, 1999). This idea of using the planning graph as a basis for computing heuristic distance metrics is further investigated in the context of state-space search in (Nguyen & Kambhampati, 2000). An interesting finding in that paper is that even when one is using state-space instead of CSP-style solution extraction, EBL can still be useful as a lazy demand-driven approach for discovering n-ary mutexes that can improve the informedness of the heuristic. Specifically, Long & Kambhampati describe a method where a limited run of Graphplan's backward search, armed with EBL/DDB is used as a pre-processing stage to explicate memos ("n-ary mutexes") which are then used to significantly improve the effectiveness of the heuristic on the state-search.

The general importance of EBL & DDB for CSP and SAT problems is well recognized. Indeed, one of the best systematic solvers for propositional satisfiability problems is RELSAT (Bayardo & Schrag, 1997), which uses EBL, DDB, and forward checking. A randomized version of RELSAT is one of the solvers supported by the BLACKBOX system (Kautz & Selman, 1999), which compiles the planning graph into a SAT encoding, and ships it to various solvers. BLACKBOX thus offers a way of indirectly comparing the Dynamic CSP and static CSP models for solving the planning graph. As discussed in Section 2.2, the main differences are that BLACKBOX needs to compile the planning graph into an extensional SAT representation. This makes it harder for BLACKBOX to exploit the results of searches in previous levels (as Graphplan does with its stored memos), and also leads to memory blowups. The latter is particularly problematic as the techniques for condensing planning graphs, such as the bi-level representation discussed in (Fox & Long, 1999; Smith & Weld, 1999) will not be effective when we compile the planning graph to SAT. On the flip side, BLACKBOX allows non-directional search, and the opportunity to exploit existing SAT solvers, rather than develop customized solvers for the planning graph. At present, it is not clear whether either of these approaches dominates the other. In my own informal experiments, I found that certain problems, such as Att-log-x, are easier to solve with non-directional search offered by BLACKBOX, while others, such as Gripper-x, are easier to solve with the Graphplan backward search. The results of the recent AIPS planning competition are also inconclusive in this respect (McDermott, 1998).

While my main rationale for focusing on dynamic CSP model of the planning graph is due to its closeness to Graphplan's backward search, Gelle (1998) argues that keeping activity constraints distinct from value constraints has several advantages in terms of modularity of the representation. In Graphplan, this advantage becomes apparent when not all activation constraints are known *a priori*, but are posted dynamically during search,. This is the case in several extensions of the Graphplan algorithm that handle conditional effects (Kambhampati et al., 1997; Anderson, Smith, & Weld, 1998; Koehler et al., 1997), and incomplete initial states (Weld, Anderson, & Smith, 1998).

Although EBL and DDB strategies try to exploit the symmetry in the search space to improve the search performance, they do not go far enough in many cases. For example, in the Gripper domain, the real difficulty is that search gets lost in the combinatorics of deciding which hand should be used to pick which ball for transfer into the next room–a decision which is completely irrelevant for the quality of the solution (or the search failures, for that matter). While EBL/DDB allow Graphplan to cut the search down a bit, allowing transfer of up to 10 balls from one room to another, they are over come beyond 10 balls. There are two possible ways of scaling further. The first is to "variablize" memos, and realize that certain types of failures would have occurred irrespective of the actual identity of the hand that is used. Variablization, also called "generalization" is a part of EBL methods (Kambhampati, 1998; Kambhampati et al., 1997). Another way of scaling up in such situations would be to recognize the symmetry inherent in the problem and abstract the resources from the search. In (Srivastava & Kambhampati, 1999), we describe this type of resource abstraction approach for Graphplan.

## 10. Conclusion and Future work

In this paper, I traced the connections between the Graphplan planning graph and CSP, and motivated the need for exploiting CSP techniques to improve the performance of Graphplan backward search. I then adapted and evaluated several CSP search techniques in the contest of Graph-plan. These included EBL, DDB, forward checking, dynamic variable ordering, sticky values, and random-restart search. My empirical studies show the EBL/DDB is particularly useful in dramatically speeding up Graphplan's backward search (by up tp 1000x in some instances). The speedups can be improved further (by up to 8x) with the addition of forward checking, dynamic variable ordering and sticky values on top of EBL/DDB. I also showed that EBL/DDB techniques are equally effective in helping Graphplan, even if random-restart search strategies are used.

A secondary contribution of this paper is a clear description of the connections between the Graphplan planning graph, and the (dynamic) constraint satisfaction problem. These connections help us understand some unique properties of the Graphplan memoization strategy, when viewed from CSP standpoint (see Section 9).

There are several possible ways of extending this work. The first would be to support the use of learned memos across problems (or when the specification of a problem changes, as is the case during replanning). Blum & Furst (1997) suggest this as a promising future direction, and the EBL framework described here makes the extension feasible. As discussed in (Kambhampati, 1998; Schiex & Verfaillie, 1993), supporting such inter-problem usage involves "contextualizing" the learned no-goods. In particular, since the soundness of memos depends only on the initial state of the problem (given that operators do not change from problem to problem), inter-problem usage of memos can be supported by tagging each learned memo with the specific initial state literals that supported that memo. Memos can then be used at the corresponding level of a new problem as long as their initial state justification holds in the new problem. The initial state justification for the memos can be computed incrementally by a procedure that first justifies the propagated mutex relations in terms of the initial state, and then justifies individual memos in terms of the justifications of the mutexes and other memos from which they are derived.

The success of EBL/DDB approaches in Graphplan is in part due to the high degree of redundancy in the planning graph structure. For example, the propositions (actions) at level $l$ in a planning graph are a superset of the propositions (actions) at level $l - 1$, the mutexes (memos) at level $l$ are a subset of the mutexes (memos) at level $l - 1$). While the EBL/DDB techniques help Graphplan exploit some of this redundancy by avoiding previous failures, the exploitation of redundancy can be pushed further. Indeed, the search that Graphplan does on a planning graph of size $l$ is almost a *re-play* of the search it did on the planning graph of size $l - 1$ (with a few additional choices). In (Zimmerman & Kambhampati, 1999), we present a complementary technique called "explanation-guided backward search" that attempts to exploit this *deja vu* property of the Graphplan's backward search. Our technique involves keeping track of an elaborate trace of the search at a level $l$ (along with the failure information), termed the "pilot explanation" for level $l$, and using the pilot explanation to guide the search at level $l - 1$. The way EBL/DDB help in this process is that they significantly reduce the size of the pilot explanations that need to be maintained. Preliminary results with this technique shows that it complements EBL/DDB and provides significant further savings in search.

### Acknowledgements

This research is supported in part by NSF young investigator award (NYI) IRI-9457634, ARPA/Rome Laboratory planning initiative grant F30602-95-C-0247, Army AASERT grant DAAH04-96-1-0247, AFOSR grant F20602-98-1-0182 and NSF grant IRI-9801676. I thank Maria Fox and Derek Long for taking the time to implement and experiment with EBL/DDB on their STAN system. I would also like to thank them, as well as Terry Zimmerman, Biplav Srivastava, Dan Weld, Avrim Blum and Steve Minton for comments on previous drafts of this paper. Special thanks are due to Dan Weld, who hosted me at University of Washington in Summer 1997, and spent time discussing the connections between CSP and Graphplan. Finally, I thank Mark Peot and David Smith for their clean Lisp implementation of the Graphplan algorithm, which served as the basis for my extensions.